\documentclass[
]{ceurart}

\usepackage{listings}
\usepackage{amssymb}
\usepackage{amsmath}
\usepackage{hyperref}
\usepackage{float}
\usepackage{graphicx}
\usepackage{subcaption}
\definecolor{codetextbkg}{RGB}{248,248,248}
\lstdefinestyle{mystyle}{
    backgroundcolor=\color{codetextbkg},   
    basicstyle=\ttfamily \scriptsize,
    numbers=none,
    breaklines=true,
    breakindent=0pt
}
\begin{document}

\copyrightyear{2024}
\copyrightclause{Copyright © 2024 for this paper by its authors. Use permitted under Creative Commons License Attribution 4.0 International (CC BY 4.0)}

\conference{EKAW 2024: EKAW 2024 Workshops, Tutorials, Posters and Demos, 24th International Conference on Knowledge Engineering and Knowledge Management (EKAW 2024), November 26-28, 2024, Amsterdam, The Netherlands.
}

\title{Evaluation of LLMs on Long-tail Entity Linking in Historical Documents}



\author[1]{Marta Boscariol}[%
orcid=0000-0001-7517-6877,
email=marta.boscariol@unito.it,
]

\address[1]{Department of Management, University of Turin, Italy
  }

\address[2]{Department of Computer Science, University of Catania, Italy
  }

\author[2]{Luana Bulla}[%
orcid=0000-0001-7116-9338,
email= luana.bulla@phd.unict.it
]

\address[3]{Department of Computer Science, University of Turin, Italy}

\author[3]{Lia Draetta}[%
orcid=0009-0004-6479-5882,
email=lia.draetta@unito.it,
]

\address[4]{ Department of Modern Languages, Literatures and Cultures, University of Bologna, Italy}

\author[4]{Beatrice Fiumanò}[%
orcid=0009-0007-6918-0803,
email=beatrice.fiumano@unibo.it,
]

\address[5]{Department of Information Engineering (DII), University of Pisa, Italy}

\author[5,7]{Emanuele Lenzi}[%
orcid=0000-0002-9159-8858,
email=emanuele.lenzi@isti.cnr.it,
]

\address[6]{Department of Mathematics and Computer Science, University of Cagliari, Italy}

\author[6]{Leonardo Piano}[%
orcid=0000-0003-1047-5491,
email=leonardo.piano@unica.it,
]

\address[7]{Institute of Information Science and Technologies (ISTI), National Research Council of Italy (CNR), Pisa, Italy}

\cormark[1]

\cortext[1]{Corresponding author.}

\begin{abstract}
  Entity Linking (EL) plays a crucial role in Natural Language Processing (NLP) applications, enabling the disambiguation of entity mentions by linking them to their corresponding entries in a reference knowledge base (KB). Thanks to their deep contextual understanding capabilities, LLMs offer a new perspective to tackle EL, promising better results than traditional methods. Despite the impressive generalization capabilities of LLMs, linking less popular, long-tail entities remains challenging as these entities are often underrepresented in training data and knowledge bases. Furthermore, the long-tail EL task is an understudied problem, and limited studies address it with LLMs. 
  In the present work, we assess the performance of two popular LLMs, GPT and LLama3, in a long-tail entity linking scenario. Using MHERCL v0.1, a manually annotated benchmark of sentences from domain-specific historical texts, we quantitatively compare the performance of LLMs in identifying and linking entities to their corresponding Wikidata entries against that of ReLiK, a state-of-the-art Entity Linking and Relation Extraction framework. Our preliminary experiments reveal that LLMs perform encouragingly well in long-tail EL, indicating that this technology can be a valuable adjunct in filling the gap between head and long-tail EL.

\end{abstract}


\maketitle

\section{Introduction}

Entity Linking (EL) is a fundamental task in Natural Language Processing (NLP) that involves the identification and disambiguation of entity mentions in text, linking them to corresponding entries in a reference Knowledge Base (KB), such as Wikipedia or Wikidata. Accurate EL enhances the understanding of text by connecting unstructured data to structured knowledge, thereby enriching the content with contextual meaning and facilitating more advanced text analytics.

The vast majority of traditional EL approaches typically rely on machine learning \cite{banerjee2020pnel, boros2020robust}, some with rule-based approaches \cite{sakor2020falcon} and others based on graph optimization \cite{klang2020hedwig}.

These methods, although effective in many cases, often struggle with ambiguous or obscure mentions, particularly when dealing with long-tail entities \cite{ilievski-etal-2018}, i.e. entities that are infrequently mentioned or have limited representation in available KBs. The scarcity of training data and the inherent diversity of long-tail entities make accurate linking a persistent challenge \cite{ilievski-etal-2018}.

The advent of Large Language Models (LLMs), such as GPT and Llama, has opened new avenues for EL. Their ability to understand complex language constructs suggests that they could enhance EL performance, especially in contexts where traditional methods falter. LLMs' extensive pre-training on several and diverse corpora allows them to handle a broad range of entities, including those that are less common or not explicitly covered in the training data \cite{li2023evaluating}. 

To investigate EL performances with long-tail entities and assess the effectiveness of LLMs in this task, the present study addresses two main research questions:
\begin{itemize}
\item How does the most reliable state-of-the-art EL tool perform with long-tail entities?
\item Are LLMs suitable for long-tail entity linking?
\end{itemize}
To do so we evaluate the performance of two LLMs (GPT and Llama), in a long-tail entity linking scenario using as benchmark MHERCL v0.1\footnote{\href{https://github.com/arianna-graciotti/historical-entity-linking/tree/main/benchmark}{https://github.com/arianna-graciotti/historical-entity-linking/tree/main/benchmark}} \cite{Graciotti2023KnowledgeEF}, a manually annotated collection of sentences from domain-specific historical texts. By comparing the performance of these LLMs against that of ReLiK \cite{orlando2024relik}, a state-of-the-art Entity Linking and Relation Extraction framework, this study aims to shed light on the potential and limitations of LLMs in handling long-tail entity linking in specialized domains. 


While long-tail entities are a fairly well-known phenomenon, relatively few researchers have addressed the long-tail EL task. For this reason, our work is part of a research area that is still largely unexplored. Additionally, this study is part of an innovative line of research that leverages LLMs for various knowledge graph-related applications. There is a clear need for further investigation into the potential roles these technologies could play across different contexts.

The present work is organized as follows: in Section \ref{Related work}, we briefly present the related work on entity linking and long-tail entities. Section \ref{Methodology} describes the methodology adopted in the experiment. In Section \ref{Experimental setup}, we introduce the experimental setup, including the dataset we use and the state-of-the-art baseline.  Sections \ref{Results} and \ref{conclusion} respectively present the results obtained and the final considerations.

\section{Related work}
\label{Related work}
Entity linking, the task of associating mentions in text with their corresponding entities in a knowledge base, has been extensively studied. Early approaches relied on heuristic-based methods \cite{nguyen2016heuristic,zheng2017collective}, and among them a prominent system is DbPedia Spotlight \cite{mendes2011dbpedia}, which automatically annotates text with DbPedia URIs, combining lexical matching techniques with context-based disambiguation. Significant advancements have been achieved by Neural Entity Linking approaches, that leverage Deep Neural Networks and Languages Models. GENRE \cite{de2020autoregressive} employs a sequence-to-sequence approach to autoregressively generate unique entity names. CHOLAN \cite{ravi2021cholan} improved EL performance by relying on a modular approach. First, it detects mentions with a BERT transformer, then it retrieves a list of candidate WikiData entities, and finally it employs another BERT model enhanced with local sentence context and Wikipedia entity descriptions to classify and link the mention to the correct entity. The aforementioned approaches involve extracting entity mentions and then linking them to a proper KB, whereas in \cite{zhang2021entqa} the author reverses this order by first retrieving candidate entities from the KB and then finding the respective mentions in the text employing a Question Answering strategy. In contrast, \textit{ReLiK} \cite{orlando2024relik} introduces a novel state-of-the-art Entity Linking and Relation Extraction system based on a Retriever-Reader architecture. Its novelty resides in using a single forward pass to link all candidate entities and extract relations, unlike previous methods that require separate passes for each candidate. This strategy permits ReLiK to achieve up to 40x faster inference compared to other methods, while maintaining strong performances.
More recently, with the advent of LLMs, several researchers have implemented EL solutions that take advantage of such technology. \textit{ChatEL} \cite{ding2024chatel} is a three-step entity linking framework where, after retrieving a set of candidate entities with \textit{BLINK} \cite{wu-etal-2020}, an LLM is prompted first to augment the entities mentions with meaningful descriptions to improve disambiguation and then to choose the correct entity. Similarly, the LLMAEL pipeline \cite{xin2024llmael} leverages LLMs as context augmenters for traditional EL models such as GENRE and BLINK, coupling their task specification capabilities with the extensive world knowledge of LLMs. Xin et al. note that this approach also enhances EL performances in long-tail scenarios, as LLMs enrich EL models with additional knowledge on low-frequency entities, facilitating entity identification and linking. Despite these advances, however, the domain of long-tail entity linking remains largely underexplored in current research, as most of the developed systems and datasets are mainly designed to capture head entities \cite{ilievski-etal-2018}.

%

\section{LLM-based Entity Linking}
\label{Methodology}
Entity linking usually involves two core tasks: (i) Entity Recognition, which detects and extracts the entity mentions from the text, and (ii) Entity Disambiguation, where the entity is correctly linked to its respective Knowledge base entry. As LLMs excel at capturing complex relationships between words, we tackled the EL problem as a sequence-to-sequence translation, jointly performing mention detection and linking with a single model interrogation. 
Formally, given a sentence \textit{S}, comprising within it a set of entities \textit{E}, where each entity is uniquely represented by a unique label (e.g Wikipedia page title), the model needs to identify each entity $e_{i}\in E$ along with its unique identifier. 
To accomplish this, we prompted the LLMs to generate a JSON-style output having as a key the textual span of the identified mention and as a value the respective Wikipedia page title. In an autoregressive fashion, the model detects the textual mentions that refer to an entity and consequently translates them into the corresponding unique identifier by probing from its knowledge. To further assist the model, we supplied an example in the prompt, thus following a one-shot approach. The employed prompt is detailed below.
\begin{lstlisting}[title=Entity Linking Prompt]
You are a powerful Entity Linking system.
Given a sentence, identify the key entities and output their exact labels as found on the corresponding Wikipedia pages.
Generate a structured JSON output, formatted as [{"Entities":{"text entity span": "Wikipedia page title"}].
Here there are some examples:
#
Sentence:"of Rameau was represented in 1735, it was a balletopera Les Indes galantes."
Output:  [{"Entities":{"Rameau":"Jean-Philippe Rameau","Les Indes galantes":"Les Indes galantes"}]
\end{lstlisting}
As an alternative identifier, it might be conceivable to exploit the QID, the unique identifier of Wikidata entities. However, through a preliminary experiment, we noticed that LLMs tended to fictionalize QIDs. In that experiment, GPT 3.5 achieved a precision of less than 1\%.
We hypothesize that this behavior is caused by the fact that QIDs mainly consist of numbers and since they don’t follow linguistic patterns, LLMs, which are trained primarily on text don't intuitively know how to generate them accurately. 
As a result, LLMs end up generating plausible-sounding yet fictional QIDs based on learned patterns. 
For the aforementioned reasons, we decided to exploit the Wikipedia page title as the unique identifier. 
For clarity, we also specify that we employ the same strategy and prompt for all the compared LLMs, which are detailed in the following section.

\section{Experimental setup}
\label{Experimental setup}
\textbf{Models.}   For our study, we harness two advanced LLMs, namely GPT 3.5 and LLama 3 \cite{dubey2024llama3herdmodels}, both in their instruct versions. Specifically, our experiments are conducted using OpenAI’s \texttt{GPT-3.5-turbo-instruct}\footnote{\href{https://platform.openai.com/docs/models/gpt-3-5-turbo}{https://platform.openai.com/docs/models/gpt-3-5-turbo}}, Meta \texttt{LLama-3-8B-instruct}\footnote{\href{https://huggingface.co/meta-llama/Meta-Llama-3-8B-Instruct}{https://huggingface.co/meta-llama/Meta-Llama-3-8B-Instruct}} and \texttt{Llama-3-70B}\footnote{\href{https://huggingface.co/meta-llama/Meta-Llama-3-70B-Instruct}{https://huggingface.co/meta-llama/Meta-Llama-3-70B-Instruct}}. GPT 3.5 Turbo is a cost-effective, cutting-edge tool ensuring deep contextual understanding, heightened accuracy and faster processing speed compared to other GPT models. LLama 3, available in configurations with 8 billion and 70 billion parameters, is an open-source, highly versatile model that offers state-of-the-art performances in a wide variety of NLP tasks, outperforming GPT models in different benchmarks and having longer context window compared to GPT-3.5 Turbo. 

\textbf{Dataset.}  The performance of LLMs is evaluated on the Musical Heritage Historical named Entities Recognition, Classification and Linking (MHERCL) benchmark. MHERCL v0.1 consists of English-language sentences extracted from the \textit{Periodicals} module of the Polifonia Textual Corpus\footnote{\href{https://github.com/polifonia-project/Polifonia-Corpus}{https://github.com/polifonia-project/Polifonia-Corpus}} (PTC), a diachronic corpus covering the domain of Musical Heritage. As part of the PTC creation, Optical Character Recognition (OCR) was leveraged to extract text from scans of historical documents. Since OCR on historical documents can be particularly challenging and prone to errors due to factors like degraded image quality or archaic fonts, this process inevitably introduced some noise into the dataset. Around 930 sentences were extracted from the PTC to create the MHERCL benchmark. Each sentence was manually annotated with EL information, including entity type and Wikidata QID. In this work, we use MHERCL v0.1.2, an expanded version of the benchmark that includes 928 sentences, 969 unique named entity mentions (NE) identified by a QID, and 59 different NE types. Table \ref{tab:stats} provides a synthetic overview of the dataset statistics. Entities that were not assigned a QID, mainly due to the lack of a corresponding Wikidata entry, were assigned a NIL label and are excluded from the dataset statistics. Since the \textit{Periodicals} module of the PTC consists of music-specialised documents published between 1823 and 1900, the MHERCL dataset features a high concentration of niche, domain-specific and historical knowledge, serving as a robust benchmark for assessing EL performances in long-tail scenarios. 

\begin{table}[h!]
  
    \begin{tabular}{cccccc}
        \toprule
        Dataset & Lang. & Sent. & Tokens & Unique NE & Types \\
        \midrule
        MHERCL v0.1.2 & EN & 928 & 28.478 & 966 & 59 \\ 
        \bottomrule
    \end{tabular}
    
    \caption{MHERCL v0.1.2 Benchmark statistics}
     \label{tab:stats}   
\end{table}

\textbf{Baseline.}  As our baseline, we leverage ReLiK, a state-of-the-art framework for Entity Linking and Relation Extraction, on the MHERCL dataset. Based on a retriever-reader architecture, ReLiK outperforms its competitors in both in-domain and out-of-domain settings, achieving better results in terms of performance, inference speed, and flexibility. ReLiK is available in three versions: small, base, and large. In our study, we leverage ReLiK-base to identify and link entity mentions within MHERCL sentences to their corresponding Wikidata entries. 
ReLiK links entities to a knowledge base other than WikiData, providing Wikipedia page IDs instead of WikiData QIDs. To map the extracted entities to their corresponding WikiData entries, we queried ReLiK's reference knowledge base, KILT \cite{petroni2020kilt}. KILT, derived from a Wikipedia dump from August 1, 2019, allowed us to retrieve WikiData IDs using either Wikipedia page IDs or entity titles.

\textbf{Evaluation.}   To evaluate and compare the performance of the selected models, we employ confusion matrix metrics such as  precision, recall, and F1-score, which are formally defined as:

\begin{equation*}
    \setlength{\jot}{10pt}
    \begin{split}     
     Precision & =\frac{TP}{TP + FP} \\
     Recall & =\frac{TP}{TP+ FN} \\
    {F1} & =\frac{2*Precision*Recall}{Precision+Recall}
    \end{split}
\end{equation*}
In computing these scores, we count a True Positive (TP) when the model’s prediction matches the ground truth, a False Positive (FP) when the model’s prediction does not match annotations in the ground truth, and a False Negative (FN) when the model is not able to identify and link the entity. Entities labeled as NIL in the MHERCL benchmark are excluded from our experiments because they generally lack corresponding Wikidata or Wikipedia entries.
We further specify that, in the case of ReLiK's results, we evaluated the match between the QID identified by ReLiK and the corresponding QID in the ground truth. In contrast, for LLMs we assessed the correct match between the predicted Wikipedia page title and the Wikipedia title retrieved using the baseline's QID.

\section{Results}
\label{Results}
This section outlines the quantitative results of our preliminary study in historical long-tail Entity Linking with Large Language Models. Table \ref{tab:comparision} highlights the comparison between LLMs against ReLiK, over the entire dataset, with no differentiation in the distribution of entities. This comparison shows that ReLiK is highly accurate, generating a low number of false positives, reaching a precision of 72.8\%. Still, it struggles to find an adequate amount of entities in such a niche domain, retrieving the 45\% of annotated entities as evidenced by the recall. Indeed, exception made for LLama3-8b, the LLMs recovered a higher number of entities, where LLama3 in the 70b configuration reached a recall score of 60.3\% exceeding the state-of-the-art ReLiK by about 15\%. Given the high recall of entities made by LLMs, we hypothesize that they could serve as entity retrievers or augment the retrieval of existing EL retrievers and we believe that this aspect may provide insights for future studies and investigations from the scientific community. 
Regardless of satisfactory recall results, the numbers dropped in the precision instance, this is because the LLMs always tended to generate some text and find fictional entities that were not annotated in the dataset, raising the number of false positives. Also, our evaluation metrics are based on exact matching between the predicted Wikipedia label and the real one, so even a single incorrectly generated character causes the prediction to be considered incorrect.
Overall, It is worth noting how the obtained results highlight the potential of LLMs in EL, as even though they are not specifically trained on the EL task, they achieved competitive results with respect to the state of the art, in an out-of-domain comparison. They even exceeded it when it comes to long-tail entity retrieval. 

\begin{table}[h!]
  \caption{Comparison between LLMs and Relik}
  \label{tab:comparision}
  \begin{tabular}{cccl}
    \toprule
    Model&Precision(\%)& Recall(\%) &F1(\%)\\
    \midrule
    ReLiK &\textbf{72.8} & 45.7&\textbf{56.1} \\
    GPT 3.5 &48.6 & 58.8& 53.2 \\
    LLama3-70b&47.3 &\textbf{60.3} & 53.0\\
    LLama3-8b & 34.9& 40.1& 37.3 \\   
   
  \bottomrule
\end{tabular}
\end{table}

Although the MHERCL benchmark is a historical, domain-specific dataset with a high number of niche and less popular entities, we conducted an additional analysis to highlight better the models' performance when varying the entities' popularity. As a measure of popularity, we leveraged the number of Wikidata triples associated with each entity, as also done in \cite{chen2023knowledge}. 

\begin{figure}[htbp]
    \centering
    
    \begin{subfigure}{0.45\textwidth}
        \centering
        \includegraphics[width=\textwidth]{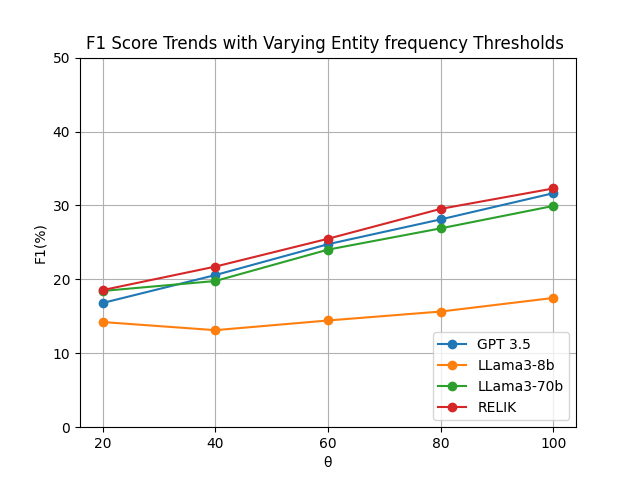}  
        
        \caption{}
        \label{fig:f1_comparision_plot}
    \end{subfigure}
    \hfill
    \begin{subfigure}{0.45\textwidth}
        \centering
        \includegraphics[width=\textwidth]{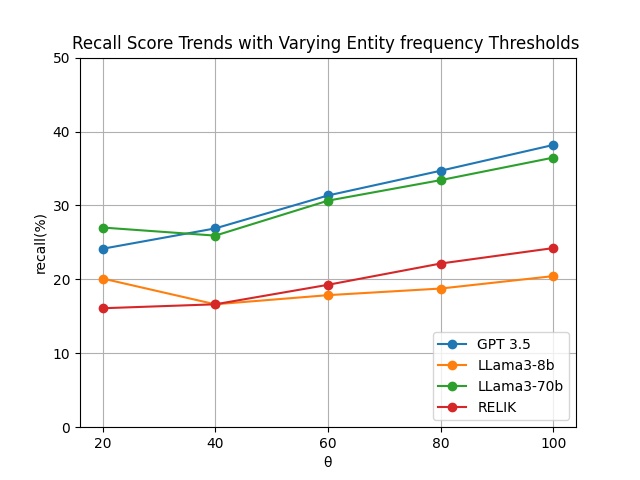}  
        \caption{}
        \label{fig:recall_comparision_plot}
    \end{subfigure}
    \caption{Measurement of the Entity Linking F1 and Recall score Across Different Entity Occurrence Thresholds for all the employed models}
    \label{fig:overall_comparision}
\end{figure}

The plot depicted in Figure \ref{fig:f1_comparision_plot}, reports the variation of the F1 score in EL at the variation of a threshold $\theta$, which has the role of differentiating real and predicted entities based on their notoriety defined by the number of Wikidata triples being associated to them. For example, a threshold $\theta$ of 20, takes into account all entities that have at most 20 triples associated with them.
The plot demonstrates that higher entity frequency thresholds $\theta$ generally lead to better EL performance for all models, likely because higher thresholds focus on more frequent, well-represented entities that are easier to disambiguate and link correctly. LLama3-70 achieves the highest f1 on par with ReLiK in linking very rare entities, having a threshold $\theta=20$. The plot in Figure \ref{fig:recall_comparision_plot} instead, highlights the recall fluctuation. GPT-3.5 and Llama 3-70b models perform better overall, with increasing recall scores as the threshold increases. ReLiK, despite being a specialized tool stays below the two largest LLMs and performs poorly with infrequent entities, obtaining the lowest recall score in the case of ~$\theta=20$. Llama 3-8b instead, demonstrates consistently lower recall and does not show significant gains with increasing entity frequency. 
The conducted analyses clearly show that the entity linking of long-tail entities is still an open challenge, as one of the most performant state-of-the-art tools was only able to retrieve $\approx$ 15\% of the annotated entities when they were less known and possessed a low frequency index. On the other hand, LLMs, at least in the larger configurations retrieved a higher number of entities, but the numbers remained unsatisfactory with a recall below 30\% and an F1 of $\approx$ 19\% referring to the less popular entities.  

\paragraph{Qualitative evaluation.}
For the sake of comprehensiveness and further interpretation of the results, we conducted a brief qualitative analysis.
Upon closer examination, we observed that a small number of entities were not correctly disambiguated by the models, due to spelling errors introduced by the OCR on the original documents. Noisy text elements are common when working with digitalized texts, especially digitalized historical documents. While human annotators were able to easily detect these mistakes and accurately identify the correct entities, many models struggled to move beyond the surface-level text. Specifically, when limited semantic context was available around the entity, both the baseline models and the LLMs struggled to accurately perform EL. For instance, given the sentence
\begin{quote}
   {\small \texttt{'Mr. Mocre is the adaptor of words to this composition, which is a tirana, arranged by Mr. Bishop.'}}
\end{quote}

none of the models were able to associate the form \textit{Mocre} with Thomas Moore (Q315346). 
On the other hand, when provided with sufficient contextual information about the entities, LLMs were more likely to identify the correct entity even in the presence of lexical errors. For example, in the sentence 
\begin{quote}
{\small \texttt{'One man may lived, who ean read the heart, and whose power was not: based upon, his own experience but if so, we may well call William Shakspeare superhuman, THenee it was that whiffe i m Rossint’s ‘Barber of Seville,’ ar Cimarosa’s ‘Seeret Marriage’}}
\end{quote}

despite the inherent difficulty due to the OCR mistakes, both GPT and LLama80b correctly associated 'Rossint' with the composer Gioachino Rossini and 'Seeret Marriage' with Domenico Cimarosa’s work \textit{The Secret Marriage}.

Nevertheless, when it comes to sentences that include less popular entities, even with appropriate context, LLMs may struggle to properly disambiguate the involved entities.
For example, when encountering the phrase 'Teatro Santo Augustino in Genoa', which should be linked to Teatro Sant'Agostino (Q19060499), both GPT 3.5 Turbo and Llama-70B incorrectly linked the entity to the more renowned Teatro Carlo Felice in Genoa.

\section{Conclusion and future works}
\label{conclusion}

In conclusion, this study highlights the potential of large language models (LLMs), such as GPT and Llama, for improving entity linking, particularly in challenging long-tail scenarios. While state-of-the-art systems like ReLiK perform well on frequent entities, LLMs show a significant advantage in identifying and linking less common, domain-specific entities, as evidenced by their higher recall scores. Despite lower precision due to occasional over-generation of entities, LLMs demonstrate the potential to recover more long-tail entities compared to ReLiK. This suggests that LLMs can serve as valuable tools in bridging the gap between frequent and infrequent entities in historical and domain-specific contexts. 
Furthermore, this study represents an early, exploratory effort to understand the efficacy of LLMs in the long-tail entity linking (EL) scenario, wherein we employed and tested relatively simple, vanilla prompt-based approaches. Thus suggesting that LLMs, even with their base unmodified form, possess inherent advantages over traditional entity-linking systems. However, there remains significant potential for further refinement. More emphasis should be placed on optimizing the balance between recall and precision. While recall is an important metric, especially for long-tail entities, precision must not be overlooked. Thus, future work should focus on developing more sophisticated prompting strategies or hybrid systems. Possible investigations include In-Context Learning (ICL) techniques, to better tailor LLMs to the task of entity linking or Knowledge Injection, to augment the LLMs' knowledge and their contextual understanding. Such methods could potentially mitigate the over-generation issue, while enhancing their accuracy in identifying and linking entities in more narrow contexts.


\begin{thebibliography}{20}
\expandafter\ifx\csname natexlab\endcsname\relax\def\natexlab#1{#1}\fi
\providecommand{\url}[1]{\texttt{#1}}
\providecommand{\href}[2]{#2}
\providecommand{\path}[1]{#1}
\providecommand{\DOIprefix}{doi:}
\providecommand{\ArXivprefix}{arXiv:}
\providecommand{\URLprefix}{URL: }
\providecommand{\Pubmedprefix}{pmid:}
\providecommand{\doi}[1]{\href{http://dx.doi.org/#1}{\path{#1}}}
\providecommand{\Pubmed}[1]{\href{pmid:#1}{\path{#1}}}
\providecommand{\bibinfo}[2]{#2}
\ifx\xfnm\relax \def\xfnm[#1]{\unskip,\space#1}\fi
\bibitem[{Banerjee et~al.(2020)Banerjee, Chaudhuri, Dubey, and Lehmann}]{banerjee2020pnel}
\bibinfo{author}{D.~Banerjee}, \bibinfo{author}{D.~Chaudhuri}, \bibinfo{author}{M.~Dubey}, \bibinfo{author}{J.~Lehmann},
\newblock \bibinfo{title}{Pnel: Pointer network based end-to-end entity linking over knowledge graphs},
\newblock in: \bibinfo{booktitle}{The Semantic Web--ISWC 2020: 19th International Semantic Web Conference, Athens, Greece, November 2--6, 2020, Proceedings, Part I 19}, \bibinfo{organization}{Springer}, \bibinfo{year}{2020}, pp. \bibinfo{pages}{21--38}.
\bibitem[{Boros et~al.(2020)Boros, Pontes, Cabrera-Diego, Hamdi, Moreno, Sid{\`e}re, and Doucet}]{boros2020robust}
\bibinfo{author}{E.~Boros}, \bibinfo{author}{E.~L. Pontes}, \bibinfo{author}{L.~A. Cabrera-Diego}, \bibinfo{author}{A.~Hamdi}, \bibinfo{author}{J.~G. Moreno}, \bibinfo{author}{N.~Sid{\`e}re}, \bibinfo{author}{A.~Doucet},
\newblock \bibinfo{title}{Robust named entity recognition and linking on historical multilingual documents},
\newblock in: \bibinfo{booktitle}{Conference and Labs of the Evaluation Forum (CLEF 2020)}, volume \bibinfo{volume}{2696}, \bibinfo{organization}{CEUR-WS Working Notes}, \bibinfo{year}{2020}, pp. \bibinfo{pages}{1--17}.
\bibitem[{Sakor et~al.(2020)Sakor, Singh, Patel, and Vidal}]{sakor2020falcon}
\bibinfo{author}{A.~Sakor}, \bibinfo{author}{K.~Singh}, \bibinfo{author}{A.~Patel}, \bibinfo{author}{M.-E. Vidal},
\newblock \bibinfo{title}{Falcon 2.0: An entity and relation linking tool over wikidata},
\newblock in: \bibinfo{booktitle}{Proceedings of the 29th ACM International Conference on Information \& Knowledge Management}, \bibinfo{year}{2020}, pp. \bibinfo{pages}{3141--3148}.
\bibitem[{Klang and Nugues(2020)}]{klang2020hedwig}
\bibinfo{author}{M.~Klang}, \bibinfo{author}{P.~Nugues},
\newblock \bibinfo{title}{Hedwig: A named entity linker},
\newblock in: \bibinfo{booktitle}{Proceedings of the Twelfth Language Resources and Evaluation Conference}, \bibinfo{year}{2020}, pp. \bibinfo{pages}{4501--4508}.
\bibitem[{Ilievski et~al.(2018)Ilievski, Vossen, and Schlobach}]{ilievski-etal-2018}
\bibinfo{author}{F.~Ilievski}, \bibinfo{author}{P.~Vossen}, \bibinfo{author}{S.~Schlobach},
\newblock \bibinfo{title}{Systematic study of long tail phenomena in entity linking},
\newblock in: \bibinfo{editor}{E.~M. Bender}, \bibinfo{editor}{L.~Derczynski}, \bibinfo{editor}{P.~Isabelle} (Eds.), \bibinfo{booktitle}{Proceedings of the 27th International Conference on Computational Linguistics}, \bibinfo{publisher}{Association for Computational Linguistics}, \bibinfo{address}{Santa Fe, New Mexico, USA}, \bibinfo{year}{2018}, pp. \bibinfo{pages}{664--674}. \URLprefix \url{https://aclanthology.org/C18-1056}.
\bibitem[{Li et~al.(2023)Li, Fang, Yang, Wang, Ye, Zhao, and Zhang}]{li2023evaluating}
\bibinfo{author}{B.~Li}, \bibinfo{author}{G.~Fang}, \bibinfo{author}{Y.~Yang}, \bibinfo{author}{Q.~Wang}, \bibinfo{author}{W.~Ye}, \bibinfo{author}{W.~Zhao}, \bibinfo{author}{S.~Zhang},
\newblock \bibinfo{title}{Evaluating chatgpt’s information extraction capabilities: An assessment of performance, explainability, calibration, and faithfulness [arxiv: 2304.11633 [cs]]},
\newblock \bibinfo{journal}{arXiv preprint arXiv:2304.11633}  (\bibinfo{year}{2023}).
\bibitem[{Graciotti(2023)}]{Graciotti2023KnowledgeEF}
\bibinfo{author}{A.~Graciotti},
\newblock \bibinfo{title}{Knowledge extraction from multilingual and historical texts for advanced question answering},
\newblock in: \bibinfo{editor}{C.~d'Amato}, \bibinfo{editor}{J.~Z. Pan} (Eds.), \bibinfo{booktitle}{Proceedings of the Doctoral Consortium at ISWC 2023 co-located with 22nd International Semantic Web Conference (ISWC 2023), Athens, Greece, November 7, 2023}, volume \bibinfo{volume}{3678} of \textit{\bibinfo{series}{{CEUR} Workshop Proceedings}}, \bibinfo{year}{2023}.
\bibitem[{Orlando et~al.(2024)Orlando, Huguet-Cabot, Barba, and Navigli}]{orlando2024relik}
\bibinfo{author}{R.~Orlando}, \bibinfo{author}{P.-L. Huguet-Cabot}, \bibinfo{author}{E.~Barba}, \bibinfo{author}{R.~Navigli},
\newblock \bibinfo{title}{Relik: Retrieve and link, fast and accurate entity linking and relation extraction on an academic budget},
\newblock \bibinfo{journal}{arXiv preprint arXiv:2408.00103}  (\bibinfo{year}{2024}).
\bibitem[{Nguyen and Ichise(2016)}]{nguyen2016heuristic}
\bibinfo{author}{K.~Nguyen}, \bibinfo{author}{R.~Ichise},
\newblock \bibinfo{title}{Heuristic-based configuration learning for linked data instance matching},
\newblock in: \bibinfo{booktitle}{Semantic Technology: 5th Joint International Conference, JIST 2015, Yichang, China, November 11-13, 2015, Revised Selected Papers 5}, \bibinfo{organization}{Springer}, \bibinfo{year}{2016}, pp. \bibinfo{pages}{56--72}.
\bibitem[{Zheng et~al.(2017)Zheng, Liu, and Liu}]{zheng2017collective}
\bibinfo{author}{G.~Zheng}, \bibinfo{author}{M.~Liu}, \bibinfo{author}{B.~Liu},
\newblock \bibinfo{title}{Collective entity linking based on dbpedia},
\newblock in: \bibinfo{booktitle}{Knowledge Graph and Semantic Computing. Language, Knowledge, and Intelligence: Second China Conference, CCKS 2017, Chengdu, China, August 26--29, 2017, Revised Selected Papers 2}, \bibinfo{organization}{Springer}, \bibinfo{year}{2017}, pp. \bibinfo{pages}{66--79}.
\bibitem[{Mendes et~al.(2011)Mendes, Jakob, Garc{\'\i}a-Silva, and Bizer}]{mendes2011dbpedia}
\bibinfo{author}{P.~N. Mendes}, \bibinfo{author}{M.~Jakob}, \bibinfo{author}{A.~Garc{\'\i}a-Silva}, \bibinfo{author}{C.~Bizer},
\newblock \bibinfo{title}{Dbpedia spotlight: shedding light on the web of documents},
\newblock in: \bibinfo{booktitle}{Proceedings of the 7th international conference on semantic systems}, \bibinfo{year}{2011}, pp. \bibinfo{pages}{1--8}.
\bibitem[{De~Cao et~al.(2020)De~Cao, Izacard, Riedel, and Petroni}]{de2020autoregressive}
\bibinfo{author}{N.~De~Cao}, \bibinfo{author}{G.~Izacard}, \bibinfo{author}{S.~Riedel}, \bibinfo{author}{F.~Petroni},
\newblock \bibinfo{title}{Autoregressive entity retrieval},
\newblock \bibinfo{journal}{arXiv preprint arXiv:2010.00904}  (\bibinfo{year}{2020}).
\bibitem[{Ravi et~al.(2021)Ravi, Singh, Mulang, Shekarpour, Hoffart, and Lehmann}]{ravi2021cholan}
\bibinfo{author}{M.~P.~K. Ravi}, \bibinfo{author}{K.~Singh}, \bibinfo{author}{I.~O. Mulang}, \bibinfo{author}{S.~Shekarpour}, \bibinfo{author}{J.~Hoffart}, \bibinfo{author}{J.~Lehmann},
\newblock \bibinfo{title}{Cholan: A modular approach for neural entity linking on wikipedia and wikidata},
\newblock \bibinfo{journal}{arXiv preprint arXiv:2101.09969}  (\bibinfo{year}{2021}).
\bibitem[{Zhang et~al.(2021)Zhang, Hua, and Stratos}]{zhang2021entqa}
\bibinfo{author}{W.~Zhang}, \bibinfo{author}{W.~Hua}, \bibinfo{author}{K.~Stratos},
\newblock \bibinfo{title}{Entqa: Entity linking as question answering},
\newblock \bibinfo{journal}{arXiv preprint arXiv:2110.02369}  (\bibinfo{year}{2021}).
\bibitem[{Ding et~al.(2024)Ding, Zeng, and Weninger}]{ding2024chatel}
\bibinfo{author}{Y.~Ding}, \bibinfo{author}{Q.~Zeng}, \bibinfo{author}{T.~Weninger},
\newblock \bibinfo{title}{Chatel: Entity linking with chatbots},
\newblock \bibinfo{journal}{arXiv preprint arXiv:2402.14858}  (\bibinfo{year}{2024}).
\bibitem[{Wu et~al.(2020)Wu, Petroni, Josifoski, Riedel, and Zettlemoyer}]{wu-etal-2020}
\bibinfo{author}{L.~Wu}, \bibinfo{author}{F.~Petroni}, \bibinfo{author}{M.~Josifoski}, \bibinfo{author}{S.~Riedel}, \bibinfo{author}{L.~Zettlemoyer},
\newblock \bibinfo{title}{Scalable zero-shot entity linking with dense entity retrieval},
\newblock in: \bibinfo{editor}{B.~Webber}, \bibinfo{editor}{T.~Cohn}, \bibinfo{editor}{Y.~He}, \bibinfo{editor}{Y.~Liu} (Eds.), \bibinfo{booktitle}{Proceedings of the 2020 Conference on Empirical Methods in Natural Language Processing (EMNLP)}, \bibinfo{publisher}{Association for Computational Linguistics}, \bibinfo{address}{Online}, \bibinfo{year}{2020}, pp. \bibinfo{pages}{6397--6407}. \URLprefix \url{https://aclanthology.org/2020.emnlp-main.519}. \DOIprefix\doi{10.18653/v1/2020.emnlp-main.519}.
\bibitem[{Xin et~al.(2024)Xin, Qi, Yao, Zhu, Zeng, Bin, Hou, and Li}]{xin2024llmael}
\bibinfo{author}{A.~Xin}, \bibinfo{author}{Y.~Qi}, \bibinfo{author}{Z.~Yao}, \bibinfo{author}{F.~Zhu}, \bibinfo{author}{K.~Zeng}, \bibinfo{author}{X.~Bin}, \bibinfo{author}{L.~Hou}, \bibinfo{author}{J.~Li}, \bibinfo{title}{Llmael: Large language models are good context augmenters for entity linking}, \bibinfo{year}{2024}. \URLprefix \url{https://arxiv.org/abs/2407.04020}. \href{http://arxiv.org/abs/2407.04020}{{\tt arXiv:2407.04020}}.
\bibitem[{{Llama Team, AI @ Meta}(2024)}]{dubey2024llama3herdmodels}
\bibinfo{author}{{Llama Team, AI @ Meta}}, \bibinfo{title}{The llama 3 herd of models}, \bibinfo{year}{2024}. \URLprefix \url{https://arxiv.org/abs/2407.21783}. \href{http://arxiv.org/abs/2407.21783}{{\tt arXiv:2407.21783}}.
\bibitem[{Petroni et~al.(2020)Petroni, Piktus, Fan, Lewis, Yazdani, De~Cao, Thorne, Jernite, Karpukhin, Maillard et~al.}]{petroni2020kilt}
\bibinfo{author}{F.~Petroni}, \bibinfo{author}{A.~Piktus}, \bibinfo{author}{A.~Fan}, \bibinfo{author}{P.~Lewis}, \bibinfo{author}{M.~Yazdani}, \bibinfo{author}{N.~De~Cao}, \bibinfo{author}{J.~Thorne}, \bibinfo{author}{Y.~Jernite}, \bibinfo{author}{V.~Karpukhin}, \bibinfo{author}{J.~Maillard}, et~al.,
\newblock \bibinfo{title}{Kilt: a benchmark for knowledge intensive language tasks},
\newblock \bibinfo{journal}{arXiv preprint arXiv:2009.02252}  (\bibinfo{year}{2020}).
\bibitem[{Chen et~al.(2023)Chen, Razniewski, and Weikum}]{chen2023knowledge}
\bibinfo{author}{L.~Chen}, \bibinfo{author}{S.~Razniewski}, \bibinfo{author}{G.~Weikum},
\newblock \bibinfo{title}{Knowledge base completion for long-tail entities},
\newblock \bibinfo{journal}{arXiv preprint arXiv:2306.17472}  (\bibinfo{year}{2023}).

\end{thebibliography}


\end{document}